\pdfoutput=1

\documentclass[11pt]{article}

\usepackage[]{naacl2021}
\usepackage[T1]{fontenc}

\usepackage[utf8]{inputenc}

\usepackage{times}
\usepackage{here}
\usepackage{latexsym}
\usepackage{booktabs}
\usepackage{bm}
\usepackage{stfloats}
\usepackage{subcaption}
\usepackage[T5,T1]{fontenc}

\DeclareTextSymbolDefault{\ohorn}{T5}
\DeclareTextSymbolDefault{\uhorn}{T5}

\usepackage{graphicx, natbib}
\usepackage{microtype}
\usepackage{amsmath}
\usepackage{amsfonts}
\usepackage{cleveref}
\usepackage{float}

\usepackage{todonotes}
\usepackage{cancel}
\usepackage{bm}
\usepackage{tikz-dependency}
\usepackage{microtype}
\usepackage{xspace}
\usepackage{amsthm}
\usepackage{amssymb}
\usepackage{comment}
\usepackage{adjustbox}
\usepackage{tipa}
\usepackage{enumitem}
\usepackage{textcomp}

\crefname{section}{\S}{\S\S}
\Crefname{section}{\S}{\S\S}
\crefname{table}{Table}{}
\crefname{figure}{Figure}{}
\crefname{algorithm}{Algorithm}{}
\crefname{equation}{eq.}{}
\crefname{appendix}{App.}{}
\crefname{prop}{Proposition}{}
\crefformat{section}{\S#2#1#3}
\usepackage{comment}

\everypar{\looseness=-1}

\newcommand{\word}[1]{\textit{#1}}

\newcommand{\UUAS}{\mathrm{UUAS}}
\newcommand{\DSpr}{\mathrm{DSpr}}
\newcommand{\BERT}{\mathrm{BERT}}
\newcommand{\GPT}{\operatorname{\mathrm{GPT-2}}}
\newcommand{\GPTbold}{\operatorname{\bm{\mathrm{GPT-2}}}}
\newcommand{\RoBERTa}{\mathrm{RoBERTa}}
\newcommand{\FastPos}{\mathrm{Fast{+}Pos}}
\newcommand{\Majority}{\mathrm{Majority}}
\newcommand{\Path}{\mathrm{Path}}

\newcommand{\FastText}{\mathrm{FastText}}

\newcommand{\defn}[1]{\textbf{#1}}

\newcommand{\red}[1]{\textcolor{red}{#1}}

\newcommand{\saveForCR}[1]{}

\def\signed #1{{\leavevmode\unskip\nobreak\hfil\penalty50\hskip2em
  \hbox{}\nobreak\hfil(#1)%
  \parfillskip=0pt \finalhyphendemerits=0 \endgraf}}

\newsavebox\mybox
\newenvironment{aquote}[1]
  {\savebox\mybox{#1}\begin{quote}}
  {\signed{\usebox\mybox}\end{quote}}

\title{Do Syntactic Probes Probe Syntax? \\ Experiments with Jabberwocky Probing}

\newcommand{\ucambridge}{\normalfont \text{1}}
\newcommand{\ethz}{\text{\normalfont 2}}
\author{Rowan Hall Maudslay$^{\ucambridge,\ethz}$~\;~~\;~Ryan Cotterell$^{\ucambridge,\ethz}$ \\
  $^{\ucambridge}$University of Cambridge~\;~~\;~$^{\ethz}$ETH Z\"{u}rich \\
  {\tt rh635@cam.ac.uk}~\;~~\;~\texttt{ryan.cotterell@inf.ethz.ch}
}

\date{}

\everypar{\looseness=-1}
\begin{document}
\everypar{\looseness=-1}

\maketitle

\begin{abstract}
Analysing whether neural language models encode linguistic information has become popular in NLP. One method of doing so, which is frequently cited to support the claim that models like $\BERT$ encode syntax, is called probing; probes are small supervised models trained to extract linguistic information from another model's output. If a probe is able to predict a particular structure, it is argued that the model whose output it is trained on must have implicitly learnt to encode it. However, drawing a generalisation about a model's linguistic knowledge about a specific phenomena based on what a probe is able to learn may be problematic: in this work, we show that semantic cues in training data means that syntactic probes do not properly isolate syntax. We generate a new corpus of semantically nonsensical but syntactically well-formed Jabberwocky sentences, which we use to evaluate two probes trained on normal data. We train the probes on several popular language models ($\BERT$, $\GPT$, and $\RoBERTa$), and find that in all settings they perform worse when evaluated on these data, for one probe by an average of $15.4$ $\UUAS$ points absolute. Although in most cases they still outperform the baselines, their lead is reduced substantially, e.g.\ by $53\%$ in the case of $\BERT$ for one probe. This begs the question: what empirical scores constitute knowing syntax?

\end{abstract}

\section{'Twas Brillig, and the Slithy Toves}
Recently, unsupervised language models like $\BERT$ \citep{devlin-etal-2019-bert} have become popular within natural language processing (NLP). 
These pre-trained sentence encoders, known affectionately as $\BERT$oids \citep{rogers-etal-2020-primer}, have pushed forward the state of the art in many NLP tasks.
Given their impressive performance, a natural question to ask is whether models like these implicitly learn to encode linguistic structures, such as part-of-speech tags or dependency trees.

There are two strains of research that investigate this question. On
one hand, \defn{stimuli-analysis} compares the relative probabilities a language model assigns to words which could fill a gap in a
cloze-style task.
This allows the experimenter to test whether neural models do well at capturing specific linguistic phenomena, such as subject--verb agreement \citep{linzen-etal-2016-assessing, gulordava-etal-2018-colorless} or negative-polarity item licensing \citep{marvin-linzen-2018-targeted, warstadt-etal-2019-investigating}. 
Another strain
of research directly analyses the neural network's representations; this is called \defn{probing}.
Probes are supervised models which attempt to predict a target linguistic structure using a model's representation as its input \citep[e.g.][]{alain2016understanding, conneau-etal-2018-cram, hupkes2018visualisation}; if the probe is able to perform the task well, then it is argued that the model has learnt to implicitly encode that structure in its representation.\footnote{Methods which analyse stimuli are also sometimes termed `probes' \citep[e.g.][]{ niven-kao-2019-probing}, but in this paper we use the term to refer specifically to supervised models.} \looseness=-1

Work from this inchoate probing literature is frequently cited to support the claim that models like $\BERT$ encode a large amount of syntactic knowledge. %
For instance, consider the two excerpts below demonstrating how a couple of syntactic probing papers have been interpreted:\footnote{\citet{jawahar-etal-2019-bert} and \citet{hewitt-manning-2019-structural} are more reserved about their claims; these examples merely show how such work is frequently interpreted, regardless of intent.}%

\begin{aquote}{\citealp{tian-etal-2020-skep}}
\textit{[The training objectives of BERT/GPT-2/XLNet] have shown great abilities to capture dependency between words and syntactic structures \citep{jawahar-etal-2019-bert}}
\end{aquote}

\begin{aquote}{\citealp{soulos-etal-2020-discovering}} \textit{Further work has found impressive degrees of syntactic structure in Transformer encodings \citep{hewitt-manning-2019-structural}} \end{aquote}

Our position in this paper is simple: %
we argue that the literature on syntactic probing is  methodologically flawed, owing to a conflation of syntax with semantics.  
We contend that no existing probing work has rigorously tested whether $\BERT$ encodes
syntax, and \textit{a fortiori} this literature should not be used to support this claim. %

To investigate whether syntactic probes actually probe syntax (or instead rely on semantics), we train two probes (\cref{sec:two_probes}) on the output representations produced by three pre-trained encoders %
on normal sentences---$\BERT$ \citep{devlin-etal-2019-bert}, $\GPT$ \citep{gpt2}, and $\RoBERTa$ \citep{roberta}. We then evaluate these probes on a novel corpus of syntactically well-formed sentences made up of pseudowords (\cref{sec:gen_jabb}), and find that their performance drops substantially in this setting: on one probe, the average $\BERT$oid $\UUAS$ is reduced by $15.4$ points, and on the other the relative advantage that $\BERT$ exhibits over a baseline drops by $53\%$. This suggests that the probes are leveraging statistical patterns in distributional semantics to aide them in the search for syntax. %
According to one of the probes, $\GPT$ falls behind a simple baseline, but in some cases the leads remains substantial, e.g.\ $20.4$ $\UUAS$ points in the case of $\BERT$. 
We use these results not to draw conclusions about any $\BERT$oids' syntactic knowledge, but instead to urge caution when drawing conclusions from probing results.
In our discussion, we contend that evaluating $\BERT$oids' syntactic knowledge requires more nuanced experimentation than simply training a syntactic probe as if it were a parser \cite{hall-maudslay-etal-2020-tale}, and call for the separation of syntax and semantics in future probing work.

\section{Syntax and Semantics}

\begin{figure}
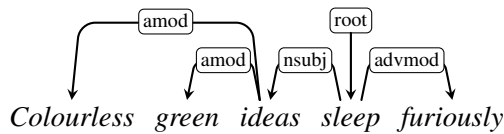

\centering
\resizebox{0.9\columnwidth}{!}{%
\begin{dependency}[] 
\begin{deptext}[column sep=0.1cm]

		\word{Colourless} \& \word{green} \& \word{ideas} \& \word{sleep} \& \word{furiously} \\
	\end{deptext}
	\depedge[edge height=6ex,edge style={thick}]{3}{1}{amod}
	\depedge[edge height=3ex,edge style={thick}]{3}{2}{amod}
	\depedge[edge height=3ex,edge style={thick}]{4}{3}{nsubj}
    \depedge[edge height=3ex,edge style={thick}]{4}{5}{advmod}
    \deproot[edge height=7.5ex, edge style={thick}]{4}{root}
\end{dependency}}

\caption{\citeauthor{chomsky1957syntactic}'s classic, albeit with the spelling corrected. %
} 
\label{fig:colourless}
\end{figure}

When investigating whether a particular model encodes syntax, those who have opted for stimuli-analysis have been careful to isolate syntactic phenomena from semantics \citep{marvin-linzen-2018-targeted, gulordava-etal-2018-colorless, goldberg}, but the same cannot be said of most syntactic probing work, which conflates the two.
To see how the two can be separated, consider the famous utterance of \newcite{chomsky1957syntactic}: 
\begin{enumerate}[label={(\arabic*)}]
	\item Colourless green ideas sleep furiously
\end{enumerate}

\noindent whose dependency parse is give in \cref{fig:colourless}. %
\citeauthor{chomsky1957syntactic}'s point is that (1) is semantically nonsensical, but syntactically well formed.

Syntactic probes are typically evaluated on real-world data, not on \citeauthor{chomsky1957syntactic}-style
sentences of (1)'s ilk. The same is true for parsers, but from a \emph{machine-learning point of view} this is not problematic, %
since the goal of a statistical parser is to parse well the data that one may encounter in the real world.
The probing literature, however, is inherently making a epistemological claim: whether $\BERT$ knows syntax.\footnote{This is not an engineering claim because the NLP engineer is unlikely to care whether $\BERT$'s representations encode syntactic structure---they just care about building reliable models that perform well on real data. An open question, however, is whether representations \emph{require} a notion of syntax to properly generalise; this is not addressed in our work.}
Indeed, we already know that $\BERT$ significantly improves the performance of statistical parsing models on real-world data \citep{zhou-zhao-2019-head}; there is no reason to develop specialist probes to reinforce that claim.
As probing consider a scientific qustion, it follows that the probing literature needs to consider syntax from \emph{a linguistic point of view} and, thus, it requires a linguistic definition of syntax.
At least in the generative tradition, it taken as definitional that grammaticality, i.e.\ syntactic well-formedness, is distinct from the meaning of the sentence. 
It is this distinction that the nascent syntactic probing literature has overlooked.

\section{Generating Jabberwocky Sentences}\label{sec:gen_jabb}

To tease apart syntax and semantics when evaluating probes, we construct a new evaluation corpus of syntactically valid English \defn{Jabberwocky sentences}, so called after \citet{jabberwocky} who wrote verse consisting in large part of pseudowords (see \cref{sec:jabberwocky}). In written language, a pseudoword is a sequence of letters which looks like a valid word in a particular language (usually determined by acceptability judgments), but which carries with it no lexical meaning. 

For our Jabberwocky corpus, we make use of the ARC Nonword Database, which contains $358,534$ monosyllabic English pseudowords \citep{rastle-etal-2002-ARC}.
We use a subset of these which were filtered out then manually validated for high plausibility by \citet{kharkwal2014taming}. 
We conjugate each of these words using hand-written rules assuming they obey the standard English morphology and graphotactics. 
This results in $1361$ word types---a total of $2377$ varieties when we annotate these regular forms with several possible fine-grained part-of-speech realisations.

\begin{figure}
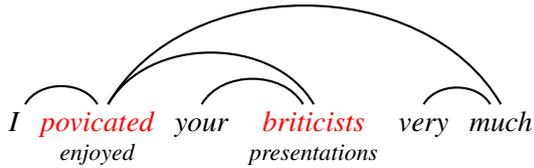

\centering
\pgfkeys{%
/depgraph/reserved/edge style/.style = {%
-, >=stealth, %
},%
}
\resizebox{0.95\columnwidth}{!}{%
\begin{dependency}[hide label, arc edge] 
\begin{deptext}[column sep=0.1cm]
		\word{I} \& \word{\red{povicated}} \& \word{your} \& \word{\red{briticists}} \& \word{very} \& \word{much} \\
		\& 	{\small \word{enjoyed}} \& 	%
		\& 	{\small \word{presentations}} \& 	%
		\& %
		\\
	\end{deptext}
	\depedge[edge height=3ex,edge style={thick}]{1}{2}{}
	\depedge[edge height=3ex,edge style={thick}]{4}{3}{}
	\depedge[edge height=4.5ex,edge style={thick}]{2}{4}{}
    \depedge[edge height=3ex,edge style={thick}]{6}{5}{}
    \depedge[edge height=6ex,edge style={thick}]{2}{6}{}
\end{dependency}} 

\caption{An unlabeled undirected parse from the EWT treebank, with Jabberwocky substitutions in red.} 
\label{fig:jabberwocky}
\end{figure}

To build sentences, we take the test portion of the English EWT Universal Dependency \citep[UD;][]{nivre-etal-2016-universal} treebank and substitute words (randomly) with our pseudowords whenever we have one available with matching fine-grained part-of-speech annotation.\footnote{More specifically, for nouns we treat elements annotated (in UD notation) with \texttt{Number=Sing} or \texttt{Number=Plur}; for verbs we treat \texttt{VerbForm=Inf},  \texttt{VerbForm=Fin | Mood=Ind | Number=Sing | Person=3 | Tense=Pres}, \texttt{VerbForm=Fin | Mood=Ind | Tense=Pres}, or \texttt{VerbForm=Part | Tense=Pres}; for adjectives and adverbs we treat \texttt{Degree=Cmp} or \texttt{Degree=Sup}, along with unmarked. These cases cover all regular forms in the EWT treebank.} Our method closely resembles \citet{kasai-frank-2019-jabberwocky}, except they do so to analyse parsers in place of syntactic probes.  %
An example of one of our Jabberwocky sentences is shown in \cref{fig:jabberwocky}, along with its unlabeled undirected parse (used by the probes) which is taken from the vanilla sentence's annotation in the treebank.

\section{Two Syntactic Probes}\label{sec:two_probes}

\saveForCR{{We note that methods of investigating model interpretability via the analysis of stimuli are also sometimes termed `probes' \citep[e.g.][]{ niven-kao-2019-probing}, but in this paper we use the term to refer specifically to supervised models.}}

A \defn{syntactic probe} is a supervised model trained to predict the syntactic structure of a sentence using representations produced by another model.  %
The main distinction between syntactic probes and dependency parsers is
one of researcher intent---probes are not meant to best the state of the art, but are a visualisation method \citep{hupkes2018visualisation}. As such, probes are typically minimally parameterised so they do not ``dig'' for information \citep[but see][]{pimentel-etal-2020-information}. If a syntactic probe performs well using a model's representations, it is argued that that model implicitly encodes syntax.

Here we briefly introduce two syntactic probes, each designed to learn the \defn{syntactic distance} between a pair of words in a sentence, which is the number of steps between them in an undirected parse tree (example in \cref{fig:jabberwocky}).
\newcite{hewitt-manning-2019-structural} first introduced syntactic distance, %
and propose %
the \defn{structural probe} as a means of identifying it; it takes a pair of embeddings %
and learns to predict the syntactic distance between them. %
An alternative to the structural probe which learns parameters for the same function is a structured perceptron dependency parser, originally introduced in \newcite{mcdonald-etal-2005-non}, and first applied to probing in \citet{hall-maudslay-etal-2020-tale}.  Here we call this the \defn{perceptron probe}.  Rather than learning syntactic distance directly, the perceptron probe instead learns to predict syntactic distances such that the minimum spanning tree that results from a sentence's predictions matches the gold standard parse tree. The difference between these probes is subtle, but they optimise for different metrics---this is reflected in our evaluation in \cref{sec:eval}.

\section{Hast Thou [Parsed] the Jabberwock?}
\label{sec:eval}

We train the probes on normal UDs, then evaluate them on Jabberwocky sentences; if the probes are really learning to extract syntax, they should perform just as well in the Jabberwocky setting.

\subsection{Experimental Setup} \label{sec:exp_des}

\begin{figure*}[ht]
    \centering
    \begin{minipage}{\columnwidth}
        \centering
        \includegraphics[width=\columnwidth,trim=0.8cm 0.6cm 2cm 1.6cm,clip]{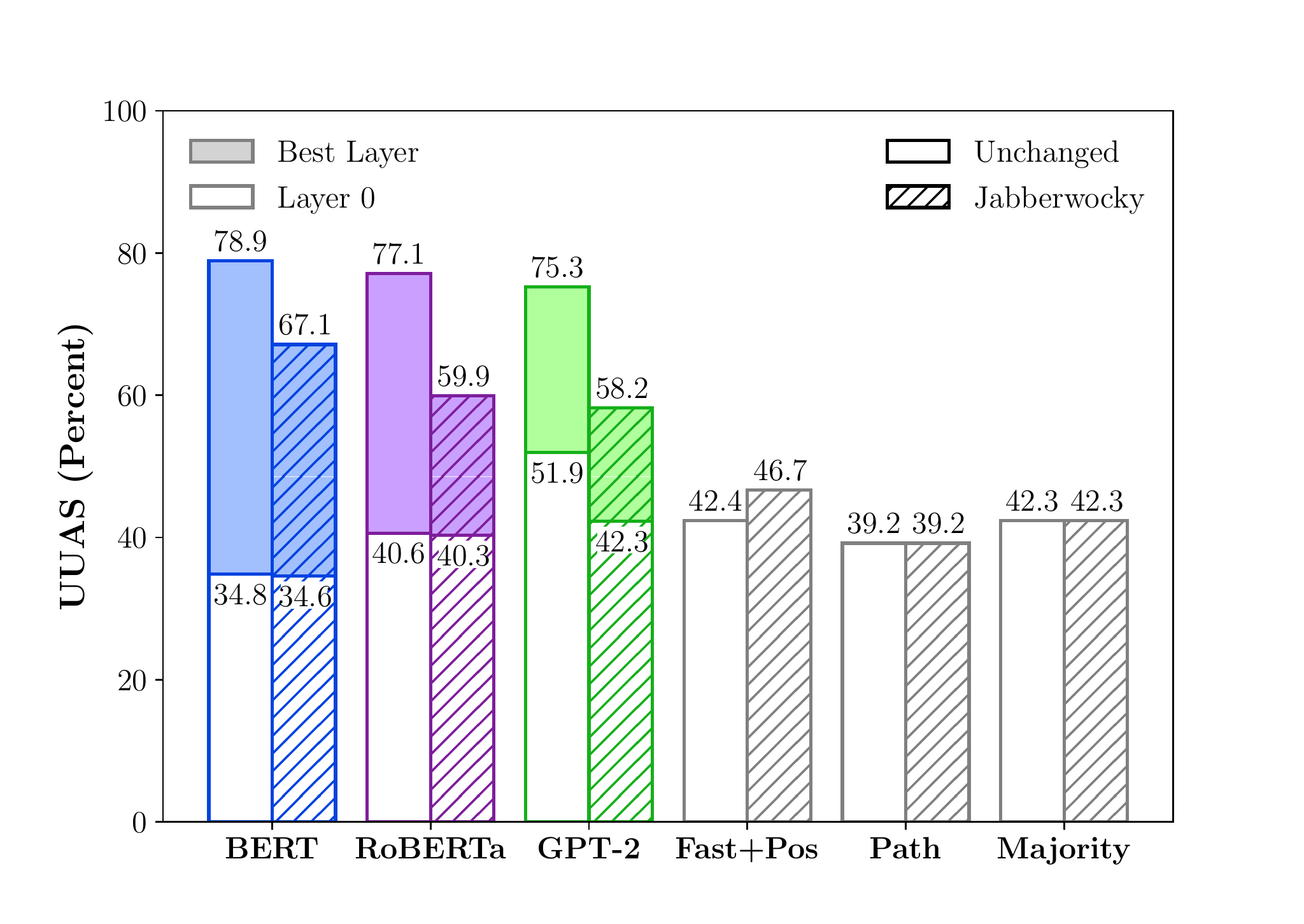}
        \subcaption[first caption.]{UUAS results from the Perceptron Probe}\label{fig:uuas}
    \end{minipage}%
    ~~~
    \begin{minipage}{\columnwidth}
        \centering
        \includegraphics[width=\columnwidth,trim=0.8cm 0.6cm 2cm 1.6cm,clip]{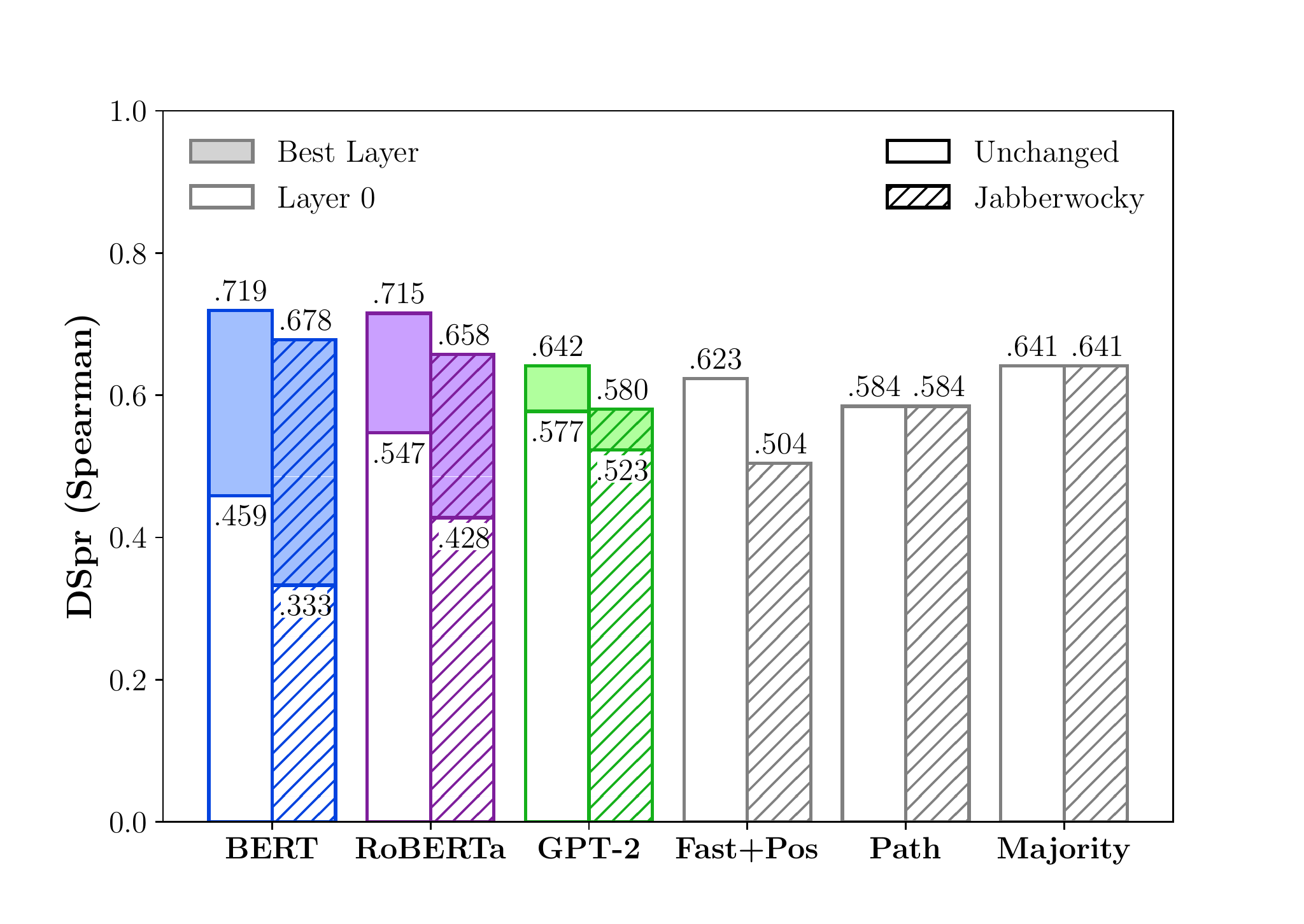}
        \subcaption[first caption.]{DSpr results from the Structural Probe}\label{fig:DSpr}
    \end{minipage}

    \caption{How the models fair when the probes are evaluated on unchanged sentences vs.\ the Jabberwocky.}
    \label{fig:normal_jabb}
\end{figure*}

\paragraph{Models to Probe} We probe three popular Transformer \citep{NIPS2017_3f5ee243} models: $\bm{\BERT}$ \citep{devlin-etal-2019-bert}, $\GPTbold$ \citep{gpt2}, and $\bm{\RoBERTa}$ \citep{roberta}. For all three we use the `large' version. We train probes on the representations at multiple layers, and choose whichever layers result in the best performance on the development set. For each Transformer model, we also train probes on the layer $0$ embeddings; we can treat these layer $0$ embeddings as baselines since they are uncontextualised, with knowledge only of a single word and where it sits in a sentence, but no knowledge of the other words. As an additional baseline representation to probe, we use $\FastText$ embeddings \citep{bojanowski-etal-2017-enriching} appended with $\BERT$ position embeddings ($\bm{\FastPos}$). We emphasise that none of these baselines can be said to encode anything about syntax (in a linguistic sense), since they are uncontextualised. Training details of these models and baselines can be found in \cref{sec:training}.

\paragraph{Additional Simple Baselines} In addition to the baseline representations which we probe, we compute two even simpler baselines, which ignore the lexical items completely. The first simply connects each word to the word next to it in a sentence ($\bm{\Path}$). %
The second returns, for a given sentence length, the tree which contains the edges occurring most frequently in the training data ($\bm{\Majority}$), which is computed as follows: first, we subdivide the training data into bins based on sentence length. For each sentence length $n$, we create an undirected graph $G_n$ with $n$ nodes, each corresponding to a different position in the sentence. The edges are weighted according to the number of times they occur in %
the training data bin which contains sentences of length $n$.
The `majority tree' of sentence length $n$ is then computed by calculating the maximum spanning tree over $G_n$, which can be done by negating the edges, then running \citeauthor{prim}'s algorithm. For $n>40$, we use the $\Path$ baseline's predictions, owing to data sparsity.

\paragraph{Metrics} As mentioned in \cref{sec:two_probes}, the probes we experiment with each optimise for subtly different aspects of syntax; %
we evaluate them on different metrics which reflect this. We evaluate the structural probe on $\DSpr$, introduced in \citet{hewitt-manning-2019-structural}---it is the Spearman correlation between the actual and predicted syntactic distances between each pair of words. We evaluate the perceptron probe using the unlabeled undirected attachment score ($\UUAS$), which is the percentage of correctly identified edges. These different metrics reflect differences in the probe designs, which are elaborated in \citet{hall-maudslay-etal-2020-tale}.

\subsection{Results}

\cref{fig:normal_jabb} shows the performance of the probes we trained, when they are evaluated on normal test data (plain) versus our specially constructed Jabberwocky data (hatched).  Recall that the test sets have identical sentence--parse structures, and differ only insofar as words in the Jabberwocky test set have been swapped for pseudowords.\footnote{This is why the $\Path$ and $\Majority$ baselines, which do not condition on the lexical items in a sentence, 
have identical scores on both datasets.} For each $\BERT$oid, the lower portion of its bars (in white) shows the performance of its layer $0$ embeddings, which are uncontextualised and thus function as additional baselines.

All the probes trained on the $\BERT$oids perform worse on the Jabberwocky data than on normal data, indicating that the probes rely in part on semantic information to make syntactic predictions. This is most pronounced with the perceptron probe: in this setting, the three $\BERT$oids' scores dropped by an average of $15.4$ $\UUAS$ points. Although they all still outperform the baselines under $\UUAS$, their advantage is less pronounced, but in some cases it remains high, e.g.\ for $\BERT$ the lead is $20.4$ points over the $\FastPos$ baseline.
With the structural probe, $\BERT$'s lead over the simple $\Majority$ baseline is reduced from $0.078$ to $0.037$ $\DSpr$, and $\RoBERTa$'s from $0.074$ to $0.017$---reductions of $53\%$ and $77\%$, respectively. $\GPT$ falls behind the baselines, and performs worse than even the simple $\Path$ predictions ($0.580$ compared to $0.584$).

\subsection{Discussion}

Is $\BERT$ still the syntactic wunderkind we had all assumed? Or do these reductions mean that these models can no longer be said to encode syntax? 
We do not use our results to make either claim. 
The reductions we have seen here may reflect a weakness of the syntactic probes %
rather than a weakness of the models themselves, per se. In order to properly give the $\BERT$oids their due, one ought train the probes on data which controls for semantic cues (e.g.\ more Jabberwocky data) in addition to evaluating them on it. 
Here, we wish only to show that existing probes leverage semantic cues to make their syntactic predictions; since they do not properly \textit{isolate} syntax, they should not be cited to support claims \textit{about} syntax.

The high performance of the baselines (which inherently contain \emph{no} syntax) is reason enough to be cautious about claims of these model's syntactic abilities. 
In general, single number metrics like these can be misleading: many correctly labeled easy dependencies may well obfuscate the mistakes being made on comparatively few hard ones, which may well be far more revealing \citep[see, for instance,][]{briscoe-carroll-2006-evaluating}.

Even if these syntactic probes achieved near perfect results on Jabberwocky data, beating the baselines by some margin, that alone would not be enough to conclude that the models encoded a deep understanding of syntax. Dependency grammarians generally parse sentences into directed graphs with labels; these probes by comparison only identify undirected unlabeled parse trees (compare Figures \ref{fig:colourless} and \ref{fig:jabberwocky} for the difference). This much-simplified version of syntax has a vastly reduced space of possible syntactic structures. %
Consider a sentence with e.g.\ $n=5$ words, for which there are only $125$ possible unlabeled undirected parse trees (by \citeauthor{cayley}'s formula, $n^{n-2}$). As the high performance of the $\Majority$ baseline indicates, these are not uniformly distributed (some parse trees are more likely than others); a probe might well use these statistical confounds to advance its syntactic predictions.
Although they remain present, biases like these are less easily exploitable in the labeled and directed case, where there are just over one billion possible parse trees to choose from.\footnote{$n\cdot n^{n-2} \cdot k^{n-1}$ where $k$ is the number of possible labels, and $k=36$ in the case of UDs  \citep{nivre-etal-2016-universal}.} 
Syntax is an incredibly rich phenomena---far more so than when it is reduced to syntactic distance.

\saveForCR{We refrain from making any comment contrasting the performance of the individual models. Our experiments do not control from such a test---each of the models was trained on different data, with a different number of parameters, and so on.}

\section{O Frabjous Day! Callooh! Callay!}

In this work, we trained two syntactic probes on a variety of $\BERT$oids, then evaluated them using Jabberwocky sentences, and showed that performance dropped substantially in this setting. This suggests that previous results from the probing literature may have overestimated $\BERT$'s syntactic abilities. %
However, in this context, we do not use the results to make any claims about $\BERT$; we contend that to make such a claim one ought train the probes on Jabberwocky sentences, which would require more psuedowords than we had available.
Instead, we 
advocate for the separation of syntax and semantics in probing. %
Future work could explore the development of artificial treebanks for use specifically for training syntactic probes, which minimise for any confounding statistical biases in the data. %
We make our Jabberwocky evaluation data and code  publicly available at \url{https://github.com/rowanhm/jabberwocky-probing}.

\bibliographystyle{acl_natbib}
\bibliography{anthology,custom}

\newpage

\appendix

\saveForCR{

\begin{figure*}[ht]
    \centering
    \begin{minipage}{\columnwidth}
        \centering
        \includegraphics[width=0.9\columnwidth,trim=0.4cm 0.0cm 1.6cm 1.1cm,clip]{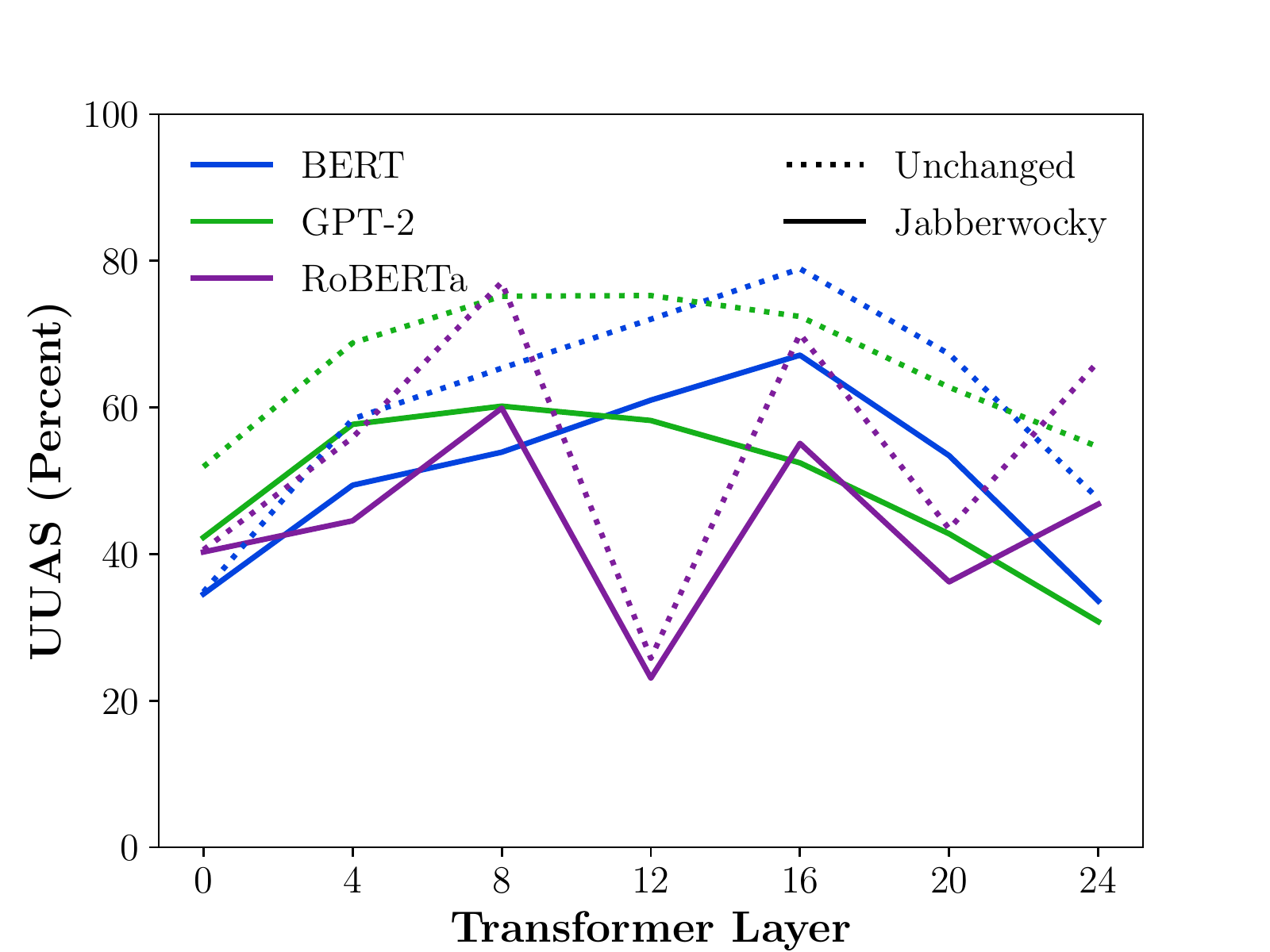}
        \subcaption[first caption.]{UUAS results from the Perceptron Probe}\label{fig:uuas}
    \end{minipage}%
    ~~~~
    \begin{minipage}{\columnwidth}
        \centering
        \includegraphics[width=0.9\columnwidth,trim=0.4cm 0.0cm 1.6cm 1.1cm,clip]{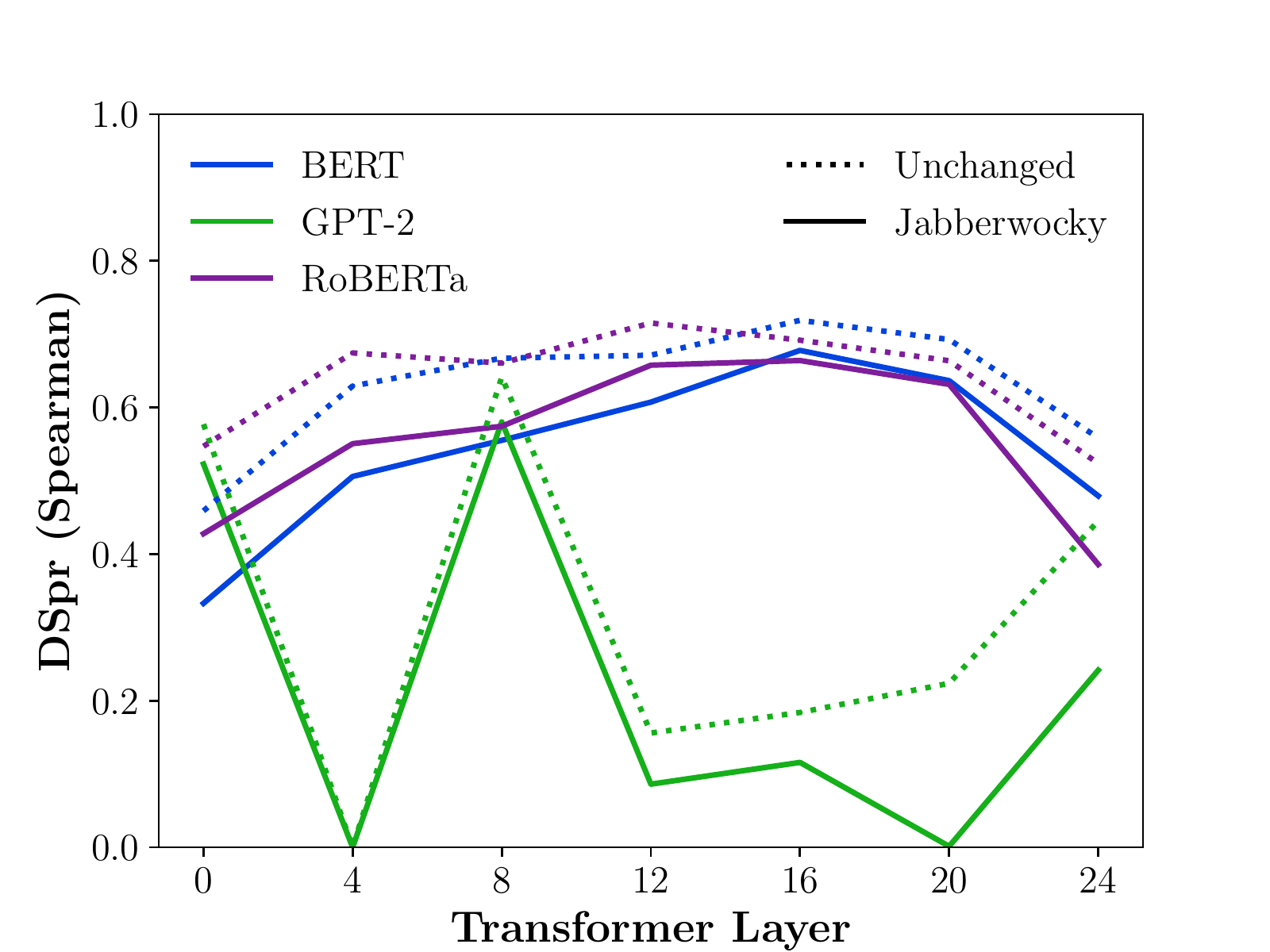}
        \subcaption[first caption.]{DSpr results from the Structural Probe}\label{fig:DSpr}
    \end{minipage}%
    \setlength{\belowcaptionskip}{-5pt}

    \caption{Probing results through the layers of the Transformer models}
    \label{fig:layers}
\end{figure*}
}

\section{The Jabberwocky} \label{sec:jabberwocky}

\begin{em}
\noindent \textquotesingle Twas brillig, and the slithy toves \\
\noindent Did gyre and gimble in the wabe; \\
\noindent All mimsy were the borogoves, \\
\noindent And the mome raths outgrabe. \\

\noindent ``Beware the Jabberwock, my son!\\
\noindent The jaws that bite, the claws that catch!\\
\noindent Beware the Jubjub bird, and shun\\
\noindent The frumious Bandersnatch!''\\

\noindent He took his vorpal sword in hand:\\
\noindent Long time the manxome foe he sought---\\
\noindent So rested he by the Tumtum tree,\\
\noindent And stood awhile in thought.\\

\noindent And as in uffish thought he stood,\\
\noindent The Jabberwock, with eyes of flame,\\
\noindent Came whiffling through the tulgey wood,\\
\noindent And burbled as it came!\\

\noindent One, two! One, two! And through and through\\
\noindent The vorpal blade went snicker-snack!\\
\noindent He left it dead, and with its head\\
\noindent He went galumphing back.\\

\noindent ``And hast thou slain the Jabberwock?\\
\noindent Come to my arms, my beamish boy!\\
\noindent O frabjous day! Callooh! Callay!''\\
\noindent He chortled in his joy.\\

\noindent \textquotesingle Twas brillig, and the slithy toves\\
\noindent Did gyre and gimble in the wabe;\\
\noindent All mimsy were the borogoves,\\
\noindent And the mome raths outgrabe.
\end{em}
\begin{flushright}{\citealp{jabberwocky}}\end{flushright}

\section{Probe Training Details} \label{sec:training}

For the $\FastPos$ baseline, we use the base model of $\BERT$, whose position embeddings are 768 dimensions, and the pretrained $\FastText$ embeddings trained on the Common Crawl (2M word variety with subword information).\footnote{The $\FastText$ embeddings are avaiable at \url{https://fasttext.cc/docs/en/english-vectors.html}} Combining the position embeddings with the 300 dimensional $\FastText$ embeddings yields embeddings with 1068 dimensions for this baseline. By comparison, the `large' version of the $\BERT$oids we train each consist of 24 layers, and produce embeddings which have 1024 dimensions.

Each $\BERT$oid we train uses a different tokenisation scheme. We need tokens which align with the tokens in the UD trees. In the case when one of the schemes does not split a word which is split in the UD trees, we merge nodes in the trees so they align. In the case where one of the schems splits a word which was not split in the UD trees, we use the first token. If the alignment is not easily fixed, we remove the sentence from the treebank. \cref{tab:data} shows the data split we are left with after sentences have been removed from the EWT UD treebank.

\begin{table}[H]
    \centering
     \begin{tabular}{lr}\toprule
     \textbf{Dataset} & \textbf{\# Sentences} \\
     \midrule
     Train & $9444$ \\
     Dev & $1400$ \\
     Test & $1398$ \\ \bottomrule
     \end{tabular}
    \caption{Sentences following removals}
    \label{tab:data}
\end{table}

To find optimimum hyperparameters, we perform a random search with 10 trials per model. When training, we used a batch size of 64 sentences, and as the optimiser we used Adam \cite{DBLP:journals/corr/KingmaB14}. 
We consider three hyperparameters: the learning rate, the rank of the probe, and Dropout \citep{JMLR:v15:srivastava14a}, over the ranges $[5\times 10^{-5}],5\times 10^{-3}]$, $[1, d]$, and $[0.1,0.8]$ respectively, 
where $d$ is the dimensionality of the input representation. 
Along with the $\FastPos$ baseline, we also perform the search on $\BERT$, $\RoBERTa$ and $\GPT$ at every fourth layer (so a total of $7$ varieties each), and choose the best layer based on loss on the development set. For each trial, we train for a maximum of 20 epochs, and use early stopping if the loss does not decrease for 15 consecutive steps. %

\end{document}